\documentclass{article}
\usepackage{spconf,amsmath,graphicx,hyperref}
\usepackage{amsmath,amssymb,graphicx,booktabs}
\usepackage{adjustbox}
\usepackage{multirow}
\usepackage{booktabs, array}
\usepackage{caption}

\usepackage[ruled,vlined]{algorithm2e}

\usepackage[table]{xcolor}

\definecolor{first}{RGB}{191, 225, 201} 
\definecolor{second}{RGB}{227, 237, 185} 
\definecolor{third}{RGB}{254, 250, 194} 

\title{Brain-HGCN: A Hyperbolic Graph Convolutional Network for Brain Functional Network Analysis}
\name{Junhao Jia$^{1,*}$, Yunyou Liu$^{1}$, Cheng Yang$^{1}$, Yifei Sun$^{2}$, Feiwei Qin$^{1,*}$\thanks{This work was supported in part by National Undergraduate Training Program for Innovation and Entrepreneurship (No.202510336076), Zhejiang Students‘ Technology and Innovation Program (No.GK250701201018 and No.GK250701201041), Fundamental Research Funds for the Provincial Universities of Zhejiang (No. GK259909299001-006), Anhui Provincial Joint Construction Key Laboratory of Intelligent Education Equipment and Technology (No. IEET202401), and the ‘Pioneer’ and ‘Leading Goose’ R\&D Program of Zhejiang (No.2025C04001).}, Changmiao Wang$^{3}$, Yong Peng$^{1}$}

\address{$^{1}$ Hangzhou Dianzi University, Hangzhou, China \\
         $^{2}$ Zhejiang University, Hangzhou, China \\
         $^{3}$ Shenzhen Research Institute of Big Data, Shenzhen, China \\
         $^{*}$ Corresponding author: 23080631@hdu.edu.cn, qinfeiwei@hdu.edu.cn
}

\begin{document}
\ninept
\maketitle
\begin{abstract}
Functional magnetic resonance imaging (fMRI) reveals complex brain functional networks with hierarchical topologies crucial for cognitive processing. Standard Euclidean Graph Neural Networks (GNNs) often struggle to represent these hierarchical structures without high distortion due to inherent spatial constraints. We propose Brain-HGCN, a geometric deep learning framework based on hyperbolic geometry, which leverages negatively curved space to model brain network hierarchy with high fidelity. Grounded in the Lorentz model, our framework employs a novel hyperbolic graph attention layer with a signed aggregation mechanism to distinctly process excitatory and inhibitory connections. Furthermore, we learn robust graph-level representations via a geometrically principled Fréchet mean for graph readout. Experiments on two large-scale fMRI datasets for psychiatric disorder classification demonstrate that Brain-HGCN significantly outperforms state-of-the-art Euclidean baselines. This work highlights the potential of hyperbolic GNNs in computational psychiatry by pioneering a new geometric paradigm for fMRI analysis.

\end{abstract}
\begin{keywords}
Hyperbolic Learning, Lorentz Model, Brain Functional Networks, Computational Psychiatry
\end{keywords}
\section{Introduction}
\label{sec:intro}
\begin{figure}[t]
\centering
\includegraphics[width=0.9\columnwidth]{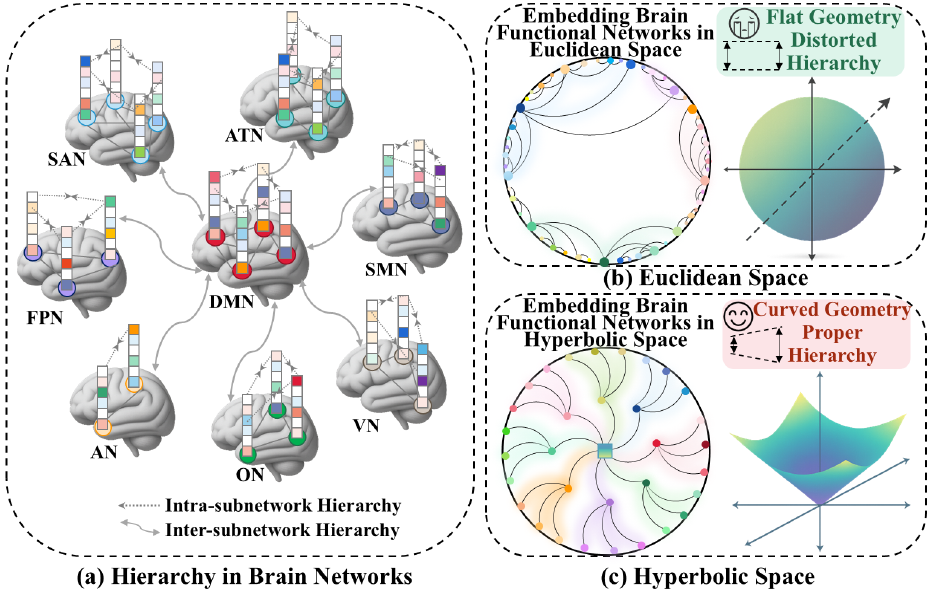}
\caption{(a) The inherent hierarchical topology of brain functional networks. (b) Euclidean embedding leads to high distortion by compressing hierarchical distances. (c) Hyperbolic embedding preserves these hierarchical relationships with high fidelity.}
\label{fig1}
\end{figure}
Psychiatric and neurodevelopmental disorders impose major global burdens, yet their neural bases remain poorly understood \cite{insel2015brain,gbd2022global}. Functional magnetic resonance imaging (fMRI) has emerged as a pivotal tool in investigating these conditions due to its non-invasive nature and high-spatiotemporal resolution, which allow for the measurement of BOLD signals and facilitate the construction of brain functional networks \cite{ogawa1990brain,fox2010clinical}. Composed of distributed brain regions and their synchronized activity, these networks exhibit a highly ordered hierarchical organization that is fundamental to the brain’s ability to process and integrate information efficiently \cite{meunier2009hierarchical,bassett2011understanding}. 

However, mainstream approaches \cite{kawahara2017brainnetcnn,kan2022brain,li2021braingnn} rely on Euclidean geometric frameworks that assume a flat data space, an assumption that is misaligned with the nonlinear topological features \cite{nickel2017poincare} inherent to hierarchical brain networks. This makes it difficult to detect subtle hierarchical abnormalities characteristic of various neurodevelopmental disorders, ultimately impeding the development of network based biomarkers for clinical diagnosis and prognosis.

To address this methodological bottleneck, hyperbolic geometry offers a natural solution for modeling hierarchical systems due to its unique exponential expansion property \cite{krioukov2010hyperbolic,Li_2025_ICCV}. Unlike Euclidean space, hyperbolic space can efficiently represent hierarchical structures in lower dimensions as its metric scale expands exponentially with distance from the center \cite{ganea2018hyperbolic}, aligning with the core concentrated periphery divergent topology of brain networks \cite{van2011rich}. Building on this, we propose a novel hyperbolic learning framework for brain functional network modeling that embeds functional connectivity data into hyperbolic space to leverage its exponential expansion property for the natural encoding of hierarchical relationships, as illustrated in Fig. \ref{fig1}. Our main contributions are:

(1) We pioneer Brain-HGCN, a geometric paradigm that embeds the brain's hierarchical topology onto the Lorentz hyperboloid to overcome Euclidean representational limitations.

(2) We architect a hyperbolic attention mechanism whose signed aggregation scheme enables geometrically principled message passing that differentiates excitatory from inhibitory pathways.

(3) We conceptualize a curvature-aware readout strategy that employs the Fréchet mean to derive distortion-free, manifold-native brain representations directly within the hyperbolic space.

\section{Preliminaries}
\label{sec:Preliminaries}

Hyperbolic geometry offers a powerful framework for modeling hierarchical data due to its intrinsic constant negative curvature. This property allows it to embed tree-like structures with minimal distortion. To leverage this geometric property in neural networks, this section introduces the core computational tools based on the Lorentz model, which lay the foundation for the model proposed in our work.

\textbf{Lorentz Model.}
The Lorentz model realizes hyperbolic space as a hyperboloid embedded in Minkowski space $\mathbb{R}^{n+1}$:
\begin{equation}
\langle \mathbf{x},\mathbf{y} \rangle_{\mathcal{L}} = -x_0y_0 + \sum_{i=1}^n x_i y_i,
\end{equation}
where the first coordinate carries negative sign.\;
The $n$-dimensional hyperbolic space with curvature $-1/K<0$ is then represented as:
\begin{equation}
\mathbb{H}^n_{K} = \{\mathbf{x}\in \mathbb{R}^{n+1} : \langle \mathbf{x},\mathbf{x}\rangle_{\mathcal{L}} = -K,\; x_0>0\},
\end{equation}
which corresponds to the upper sheet of a two-sheeted hyperboloid.\;
This embedding provides a smooth Riemannian manifold of constant negative curvature $-1/K$.\;
The geodesic distance between two points $\mathbf{x},\mathbf{y}\in \mathbb{H}^n_{K}$ is:
\begin{equation}
d_{K}(\mathbf{x},\mathbf{y}) = \operatorname{arcosh}\!\Big(-\frac{\langle \mathbf{x},\mathbf{y}\rangle_{\mathcal{L}}}{K}\Big).
\end{equation}
This distance reflects the exponential expansion property of hyperbolic space, making it suited for encoding hierarchical relations.

\textbf{Tangent Space.}
At each point $\mathbf{x}\in \mathbb{H}^n_{K}$, the tangent space is the linear space orthogonal to $\mathbf{x}$ under the Lorentzian product:
\begin{equation}
T_{\mathbf{x}}\mathbb{H}^n_{K} = \{\mathbf{v}\in \mathbb{R}^{n+1} : \langle \mathbf{v},\mathbf{x}\rangle_{\mathcal{L}}=0\}.
\end{equation}
The tangent space provides a local flat approximation of the curved manifold, enabling us to carry out neural operations such as linear transformations and aggregations in a consistent way.

\textbf{Exponential and Logarithmic Maps.}
The exponential map projects a tangent vector back to the manifold. For $\mathbf{v}\in T_{\mathbf{x}}\mathbb{H}^n_{K}$,
\begin{equation}
\exp_{\mathbf{x}}^{K}(\mathbf{v}) =
\cosh\!\Big(\frac{\|\mathbf{v}\|_{\mathcal{L}}}{\sqrt{K}}\Big)\,\mathbf{x}
\;+\;
\sqrt{K}\,\sinh\!\Big(\frac{\|\mathbf{v}\|_{\mathcal{L}}}{\sqrt{K}}\Big)\,\frac{\mathbf{v}}{\|\mathbf{v}\|_{\mathcal{L}}},
\end{equation}
where $\|\mathbf{v}\|_{\mathcal{L}} = \sqrt{\langle \mathbf{v},\mathbf{v}\rangle_{\mathcal{L}}}$.\;
Intuitively, $\exp_{\mathbf{x}}^{K}$ moves a geodesic distance $\|\mathbf{v}\|_{\mathcal{L}}/\sqrt{K}$ from $\mathbf{x}$ along the geodesic in the direction of $\mathbf{v}$.

The logarithmic map is the inverse, mapping a point $\mathbf{y}\in \mathbb{H}^n_{K}$ back to the tangent space at $\mathbf{x}$:
\begin{equation}
\log_{\mathbf{x}}^{K}(\mathbf{y}) =
\frac{\operatorname{arcosh}\!\big(-\tfrac{\langle \mathbf{x},\mathbf{y}\rangle_{\mathcal{L}}}{K}\big)}
     {\sqrt{\big(-\tfrac{\langle \mathbf{x},\mathbf{y}\rangle_{\mathcal{L}}}{K}\big)^2-1}}
\left(\mathbf{y} + \tfrac{\langle \mathbf{x},\mathbf{y}\rangle_{\mathcal{L}}}{K} \, \mathbf{x}\right).
\end{equation}
This operation returns the tangent vector at $\mathbf{x}$ that points towards $\mathbf{y}$, whose Lorentzian norm satisfies $\|\log_{\mathbf{x}}^{K}(\mathbf{y})\|_{\mathcal{L}}=\sqrt{K}\,d_{K}(\mathbf{x},\mathbf{y})$.

\section{Methodology}
\label{sec:method}

\begin{figure*}[t]
\centering
\includegraphics[width=0.8\textwidth]{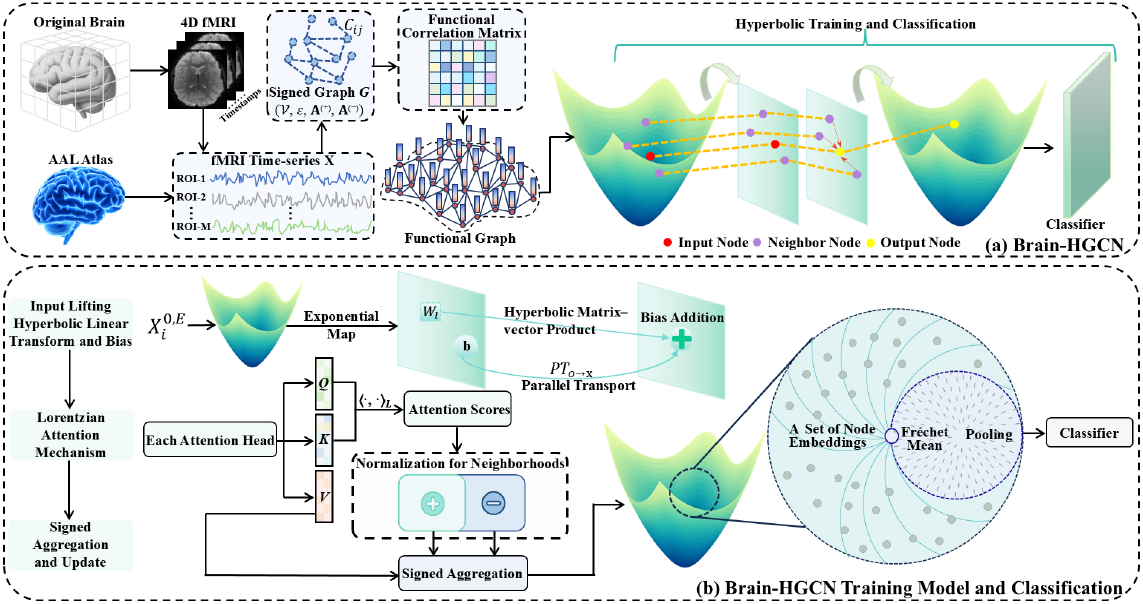}
\caption{The overview of our proposed Brain-HGCN framework. (a) Illustrates the process of constructing a signed functional graph from fMRI data for end-to-end classification. (b) The core model for Brain-HGCN training and classification, featuring a Lorentzian attention with signed aggregation for message passing, and a Fréchet mean readout for intrinsic graph pooling and classification.}
\label{fig2}
\end{figure*}

As illustrated in Fig.~\ref{fig2}, we propose the Brain-HGCN framework, which embeds fMRI data into hyperbolic space and exploits this space’s intrinsic exponential expansion to naturally capture and encode the hierarchical organizational principles of the brain.

\subsection{Subject-wise Brain Network Construction}
For each subject, we build a weighted, undirected, signed graph 
$\mathcal{G} = (\mathcal{V}, \mathcal{E}, \mathbf{A}^{(+)}, \mathbf{A}^{(-)})$.
The nodes $\mathcal{V}$ represent Regions of Interest (ROIs) from the AAL-116 atlas. 
The initial feature for each node $v_i$ is its corresponding ROI-averaged time series, 
$\mathbf{x}_i^{0,E} \in \mathbb{R}^T$, where $T$ is the number of time points. 
Edges $\mathcal{E}$ are weighted by the Pearson correlation $C_{ij}$ between nodal time series. 
To preserve signed coupling, we construct a sparse, symmetric graph by retaining the top-$k$ positive and negative connections for each node:
\begin{equation}
A^{(+)}_{ij}=\max(C_{ij},0),\qquad A^{(-)}_{ij}=\max(-C_{ij},0).
\end{equation}

\subsection{Hyperbolic Neural Network Primitives}
We instantiate the model on the {hyperboloid $\mathbb{H}^{d}_{K}$ with constant negative curvature $-1/K<0$. Each layer learns its own curvature $K_\ell>0$. Let $\mathbf{o}=(\sqrt{K_0},0,\ldots,0)$ denote the origin. We use the exponential/logarithmic maps $\exp^{K}{(\cdot)}(\cdot)$ and $\log^{K}{(\cdot)}(\cdot)$. Superscripts $E$/$H$ distinguish Euclidean and hyperbolic quantities.

\textbf{Input lifting.} Euclidean node features are lifted to the manifold via the exponential map at the origin:
\begin{equation}
\mathbf{x}^{0,H}_i = \exp^{K_0}_{\mathbf{o}}\!\big((0,\mathbf{x}^{0,E}_i)\big).
\end{equation}

\textbf{Hyperbolic linear transform and bias.} Let $\mathbf{W}_\ell$ be a weight matrix acting in the origin tangent space. We define the hyperbolic matrix--vector product and bias addition by:

\begin{equation}
\begin{aligned}
\mathbf{W}_\ell \otimes_{K_{\ell-1}} \mathbf{x}^{\ell-1,H}_i 
&= \exp^{K_{\ell-1}}_{\mathbf{o}}\!\Big(\mathbf{W}_\ell\,\log^{K_{\ell-1}}_{\mathbf{o}}(\mathbf{x}^{\ell-1,H}_i)\Big), \\
\mathbf{x}\ \oplus_{K}\ \mathbf{b} 
&= \exp^{K}_{\mathbf{x}}\!\Big(\mathrm{PT}^{K}_{\mathbf{o}\to \mathbf{x}}(\mathbf{b})\Big).
\end{aligned}
\end{equation}
where $\mathbf{b}_\ell\in T_{o}\mathbb{H}^{d}_{K_{\ell-1}}$ is a bias in the origin tangent space and $\mathrm{PT}^{K}_{o\to \mathbf{x}}$ denotes parallel transport to the center point’s tangent space. The transformed representation is:

\begin{equation}
\mathbf{h}^{\ell,H}_i \;=\; \big(\mathbf{W}_\ell \otimes_{K_{\ell-1}} \mathbf{x}^{\ell-1,H}_i\big)\ \oplus_{K_{\ell-1}}\ \mathbf{b}_\ell.
\end{equation}

\subsection{Hyperbolic Attention with Signed Aggregation}

We introduce a hyperbolic attention layer designed for the signed topology of brain networks. Let $\mathbf{x}^{\ell-1,H}_i$ be the feature of node $i$ from the previous layer. First, we apply a hyperbolic linear transformation and add a bias to obtain an intermediate representation:
\begin{equation}
\mathbf{h}^{\ell,H}_i \;=\; \big(\mathbf{W}_\ell \otimes_{K_{\ell-1}} \mathbf{x}^{\ell-1,H}_i\big)\ \oplus_{K_{\ell-1}}\ \mathbf{b}_\ell.
\end{equation}

\textbf{Lorentzian Attention Mechanism.}
We adapt multi-head attention to the Lorentz model. For each head $m$, queries ($\mathbf{q}$), keys ($\mathbf{k}$), and values ($\mathbf{v}$) are generated via linear maps and attention scores are computed using the Lorentzian inner product $\langle \cdot,\cdot\rangle_{\mathcal L}$, which naturally arises from the geometry of the Lorentz model:
\begin{equation}
\label{eq:lorentz-sdp}
s_{ij}^{(m)} \;=\; \frac{ \langle \mathbf q_i^{(m)},\, \mathbf k_j^{(m)} \rangle_{\mathcal L} }
{ \sqrt{d}\;\tau_\ell },
\quad \text{where} \quad
\tau_\ell \;=\; \tau_0/\sqrt{K_{\ell-1}}.
\end{equation}
To handle signed edges, we normalize scores separately over excitatory ($+$) and inhibitory ($-$) neighborhoods:
\begin{equation}
\begin{split}
w_{ij}^{(m,+)} &= \mathrm{softmax}_{j\in\mathcal N_i^{(+)}}\!\big(s_{ij}^{(m)}\big), \\
w_{ij}^{(m,-)} &= \mathrm{softmax}_{j\in\mathcal N_i^{(-)}}\!\big(s_{ij}^{(m)}\big).
\end{split}
\end{equation}

\textbf{Signed Aggregation and Update.}
The aggregation is performed in the tangent space of the central node $\mathbf{h}^{\ell,H}_i$. We pull from positive neighbors and push from negative ones:
\begin{equation}
\label{eq:signed-center-agg}
\begin{aligned}
\Delta_i^{(m)} \;=\;&
\sum_{j\in\mathcal N_i^{(+)}} w_{ij}^{(m,+)} \,\log^{K_{\ell-1}}_{\mathbf h^{\ell,H}_i}(\mathbf v_j^{(m)}) \\
&-\!\sum_{j\in\mathcal N_i^{(-)}} w_{ij}^{(m,-)} \,\log^{K_{\ell-1}}_{\mathbf h^{\ell,H}_i}(\mathbf v_j^{(m)}).
\end{aligned}
\end{equation}
The outputs are averaged and projected back to the manifold:
\begin{equation}
\mathbf y^{\ell,H}_i
\;=\;
\exp^{K_{\ell-1}}_{\mathbf h^{\ell,H}_i}\!\Big(\frac{1}{H}\sum_{m=1}^{H}\Delta_i^{(m)}\Big).
\end{equation}
Finally, we apply a Euclidean non-linearity $\sigma$ in the origin's tangent space and map the result back to the manifold under the new layer curvature $K_\ell$ to obtain the final node representation for this layer:
\begin{equation}
\mathbf{x}^{\ell,H}_i\;=\;\exp^{K_\ell}_{\mathbf{o}}\!\Big(\sigma\big(\log^{K_{\ell-1}}_{\mathbf{o}}(\mathbf{y}^{\ell,H}_i)\big)\Big).
\end{equation}

\subsection{Intrinsic Graph Readout and Classification}
\label{sec:readout-objective}

The final hyperbolic layer yields node embeddings $\{\mathbf{x}^{L,H}_i\}_{i=1}^{N}$. For graph-level classification, our intrinsic readout computes the Fréchet mean and pools in its tangent space.

\textbf{Fréchet Mean as Geometric Center.}
The Fréchet mean, $\boldsymbol{\mu}_{\mathcal{G}}$ generalizes the Euclidean centroid to Riemannian manifolds, defined as the point minimizing the sum of squared geodesic distances \cite{jia2025geodesic}:
\begin{equation}
\label{eq:frechet}
\boldsymbol{\mu}_{\mathcal{G}}
\;=\;
\arg\min_{x \in \mathbb{H}^d_{K_L}}
\frac{1}{N}\sum_{i=1}^{N} d_{K_L}\!\big(x, \mathbf{x}^{L,H}_i\big)^2.
\end{equation}
On Hadamard manifolds, this mean is unique and can be found by Karcher flow:
\begin{equation}
\label{eq:karcher-flow}
\begin{aligned}
\mathbf{v}^{(t)} &= \frac{1}{N}\sum_{i=1}^{N}
\log^{K_L}_{\boldsymbol{\mu}^{(t)}}\!\big(\mathbf{x}^{L,H}_i\big), \\
\boldsymbol{\mu}^{(t+1)} &=
\exp^{K_L}_{\boldsymbol{\mu}^{(t)}}\!\big(\eta\,\mathbf{v}^{(t)}\big).
\end{aligned}
\end{equation}

\textbf{Pooling and Classification.}
With $\boldsymbol{\mu}_{\mathcal{G}}$ as the anchor, we map nodes to $T_{\boldsymbol{\mu}_{\mathcal{G}}}$ and average:
\begin{equation}
\label{eq:readout-mean}
\mathbf{z}_{\mathcal{G}}
\;=\;
\frac{1}{N}\sum_{i=1}^{N}
\log^{K_L}_{\boldsymbol{\mu}_{\mathcal{G}}}\!\big(\mathbf{x}^{L,H}_i\big).
\end{equation}
The Euclidean vector $\mathbf{z}_{\mathcal{G}}$ is fed to a classifier and we train end-to-end with cross-entropy loss.

\begin{table*}[!t]
\caption{Performance Comparison on ADHD-200 and ABIDE datasets. All results are reported as mean$\pm$std over a 10-fold cross-validation. The top three results in each column are highlighted in \colorbox{first}{first}, \colorbox{second}{second}, and \colorbox{third}{third} place colors, respectively.}
\centering
\fontsize{9pt}{11pt}\selectfont
\setlength{\tabcolsep}{7pt}
\renewcommand{\arraystretch}{0.8}
\begin{tabular}{
>{\raggedright\arraybackslash}p{4.0cm}
>{\centering\arraybackslash}p{1.2cm}
>{\centering\arraybackslash}p{1.2cm}
>{\centering\arraybackslash}p{1.2cm}
>{\centering\arraybackslash}p{1.2cm}
>{\centering\arraybackslash}p{1.2cm}
>{\centering\arraybackslash}p{1.2cm}
>{\centering\arraybackslash}p{1.2cm}
>{\centering\arraybackslash}p{1.2cm}
}
\toprule
\multicolumn{1}{c}{\multirow{2}{*}[-0.3ex]{\textbf{Method}}} & \multicolumn{4}{c}{ADHD-200 (ADHD)\cite{bellec2017neuro}} & \multicolumn{4}{c}{ABIDE (ASD)\cite{di2014autism}} \\
\cmidrule(lr){2-5} \cmidrule(lr){6-9}
& ACC (\%) & SEN (\%) & SPE (\%) & AUC (\%) & ACC (\%) & SEN (\%) & SPE (\%) & AUC (\%) \\
\midrule
\multicolumn{9}{l}{\textbf{CNN-based Method}} \\
\midrule
VGG16 (ICLR'15)\cite{simonyan2014very}                         & 64.6$\pm$3.8 & 59.7$\pm$5.3 & 70.3$\pm$3.1 & 65.0$\pm$3.1 & 71.3$\pm$3.2 & 73.0$\pm$4.2 & 69.1$\pm$3.9 & 71.0$\pm$2.9 \\
ResNet-50 (CVPR'16)\cite{he2016deep}                           & 68.5$\pm$3.4 & 57.7$\pm$4.8 & 78.1$\pm$3.6 & 67.9$\pm$3.0 & 72.2$\pm$2.9 & 68.0$\pm$3.3 & 75.6$\pm$2.6 & 71.8$\pm$2.1 \\
BrainNetCNN (Neuro'17)\cite{kawahara2017brainnetcnn} & 68.0$\pm$3.3 & 65.8$\pm$8.3 & 70.4$\pm$5.3 & 68.1$\pm$4.9 & 67.3$\pm$2.6 & 63.3$\pm$9.4 & 70.5$\pm$8.7 & 66.9$\pm$6.4 \\
MMAFN (DLCV'25)\cite{jia2025mmafn}                                                 & 77.6$\pm$2.1 & 75.2$\pm$2.7 & 73.5$\pm$2.1 & 77.4$\pm$2.3 & 80.4$\pm$1.6 & 79.2$\pm$2.4 & 77.5$\pm$2.1 & 80.1$\pm$1.4 \\
\midrule
\multicolumn{9}{l}{\textbf{Transformer-based Method}} \\
\midrule
BNT (NeurIPS'22)\cite{kan2022brain}                  & 72.8$\pm$2.9 & 70.9$\pm$4.3 & 73.9$\pm$3.6 & 72.4$\pm$2.8 & 75.9$\pm$1.6 & 72.9$\pm$5.3 & 69.8$\pm$6.6 & 71.3$\pm$4.2 \\
MCPATS (JBHI'24)\cite{jiang2024multi}                & 74.0$\pm$3.3 & 64.3$\pm$4.7 & 80.4$\pm$2.9 & 72.3$\pm$3.8 & 76.0$\pm$2.6 & 73.2$\pm$3.6 & 78.7$\pm$2.9 & 76.0$\pm$2.3 \\
GBT (MICCAI'24)\cite{peng2024gbt}                    & 70.4$\pm$3.6 & 68.1$\pm$4.3 & 71.5$\pm$3.9 & 69.8$\pm$2.9 & 78.0$\pm$6.6 & 79.5$\pm$9.5 & 77.3$\pm$4.0 & 78.4$\pm$5.2 \\
RTGMFF (BIBM'25)\cite{jia2025rtgmff}                                        & \cellcolor{third}{80.7$\pm$2.5} & \cellcolor{second}{79.5$\pm$3.0} & \cellcolor{second}{81.3$\pm$2.8} & \cellcolor{third}{80.4$\pm$2.1} & \cellcolor{second}{86.4$\pm$1.9} & \cellcolor{third}{84.5$\pm$2.7} & \cellcolor{second}{87.5$\pm$2.3} & \cellcolor{third}{86.0$\pm$1.8} \\
\midrule
\multicolumn{9}{l}{\textbf{GNN-based Method}} \\
\midrule
BrainGNN (MIA'21)\cite{li2021braingnn}               & 64.7$\pm$3.8 & 67.9$\pm$3.5 & 62.7$\pm$4.1 & 65.3$\pm$2.7 & 69.3$\pm$3.9 & 66.8$\pm$3.3 & 72.7$\pm$3.4 & 69.8$\pm$2.4 \\
BrainGB (TMI'22)\cite{cui2022braingb}                & 65.8$\pm$2.6 & 61.3$\pm$4.9 & 70.3$\pm$3.8 & 65.8$\pm$3.1 & 63.2$\pm$2.0 & 63.8$\pm$8.1 & 60.1$\pm$6.9 & 62.0$\pm$5.3 \\
A-GCL (MIA'23)\cite{zhang2023gcl}                    & 77.8$\pm$4.4 & 76.4$\pm$4.7 & 79.4$\pm$5.3 & 77.9$\pm$3.5 & 82.9$\pm$2.1 & 82.4$\pm$2.6 & \cellcolor{third}{83.7$\pm$1.9} & 83.1$\pm$1.6 \\
KMGCN (MIA'25)\cite{zeng2025knowledge}               & 75.2$\pm$2.6 & 72.8$\pm$3.9 & 77.0$\pm$2.6 & 74.2$\pm$3.9 & \cellcolor{third}{84.7$\pm$1.3} & 83.6$\pm$1.7 & 81.3$\pm$1.4 & 82.4$\pm$1.1 \\
MM-GTUNets (TMI'25) \cite{cai2025mm}                                  & \cellcolor{second}{81.7$\pm$1.6} & \cellcolor{third}{78.8$\pm$2.9} & \cellcolor{third}{81.1$\pm$1.9} & \cellcolor{second}{88.7$\pm$1.5} & 83.1$\pm$1.7 & \cellcolor{second}{84.6$\pm$2.1} & 82.3$\pm$1.6 & \cellcolor{second}{87.4$\pm$1.5} \\
\midrule
\textbf{Brain-HGCN (Ours)}                            & \cellcolor{first}{83.6$\pm$1.3} & \cellcolor{first}{80.8$\pm$2.7} & \cellcolor{first}{85.8$\pm$1.6} & \cellcolor{first}{90.7$\pm$1.2} & \cellcolor{first}{87.3$\pm$1.1} & \cellcolor{first}{89.6$\pm$1.2} & \cellcolor{first}{84.3$\pm$0.9} & \cellcolor{first}{91.4$\pm$0.7} \\
\bottomrule
\end{tabular}
\label{tab1}
\end{table*}

\begin{table*}[!t]
\caption{Ablation study on the ADHD-200 and ABIDE datasets. All results are reported as mean$\pm$std over a 10-fold cross-validation. The top three results in each column are highlighted in \colorbox{first}{first}, \colorbox{second}{second}, and \colorbox{third}{third} place colors, respectively.}
\centering
\fontsize{9pt}{11pt}\selectfont
\setlength{\tabcolsep}{7pt}
\renewcommand{\arraystretch}{1.1}
\begin{tabular}{lcccccccc}
\toprule
\multicolumn{1}{c}{\multirow{2}{*}[-0.3ex]{\textbf{Method}}} & \multicolumn{4}{c}{ADHD-200 (ADHD)\cite{bellec2017neuro}} & \multicolumn{4}{c}{ABIDE (ASD)\cite{di2014autism}} \\
\cmidrule(lr){2-5} \cmidrule(lr){6-9}
& ACC (\%) & SEN (\%) & SPE (\%) & AUC (\%) & ACC (\%) & SEN (\%) & SPE (\%) & AUC (\%) \\
\midrule
w/o Hyperbolic Geometry & 78.5$\pm$2.1 & 75.1$\pm$3.5 & 80.3$\pm$2.8 & 84.1$\pm$1.9 & 82.1$\pm$1.8 & 83.2$\pm$2.5 & 80.5$\pm$2.1 & 85.3$\pm$1.5 \\
w/o Fréchet Mean Readout & \cellcolor{second}{82.0$\pm$1.5} & \cellcolor{second}{78.9$\pm$2.9} & \cellcolor{second}{84.1$\pm$1.9} & \cellcolor{second}{88.5$\pm$1.4} & \cellcolor{second}{85.9$\pm$1.3} & \cellcolor{second}{87.5$\pm$1.8} & \cellcolor{third}{83.1$\pm$1.2} & \cellcolor{second}{89.8$\pm$1.0} \\
w/o Lorentzian Attention & \cellcolor{third}{81.3$\pm$1.8} & \cellcolor{third}{78.2$\pm$3.1} & \cellcolor{third}{83.5$\pm$2.2} & \cellcolor{third}{87.9$\pm$1.6} & \cellcolor{third}{85.2$\pm$1.5} & \cellcolor{third}{86.9$\pm$2.0} & 82.8$\pm$1.4 & \cellcolor{third}{89.1$\pm$1.2} \\
w/o Signed Aggregation & 80.2$\pm$1.9 & 76.9$\pm$3.3 & 82.8$\pm$2.5 & 86.5$\pm$1.8 & 84.5$\pm$1.6 & 85.1$\pm$2.2 & \cellcolor{second}{83.7$\pm$1.3} & 88.2$\pm$1.3 \\
\midrule
\textbf{Brain-HGCN (Ours)} & \cellcolor{first}{83.6$\pm$1.3} & \cellcolor{first}{80.8$\pm$2.7} & \cellcolor{first}{85.8$\pm$1.6} & \cellcolor{first}{90.7$\pm$1.2} & \cellcolor{first}{87.3$\pm$1.1} & \cellcolor{first}{89.6$\pm$1.2} & \cellcolor{first}{84.3$\pm$0.9} & \cellcolor{first}{91.4$\pm$0.7} \\
\bottomrule
\end{tabular}
\label{tab:ablation}
\end{table*}

\section{Experiments}
\label{sec:Experiments}
\subsection{Experimental Setup}
We evaluate Brain-HGCN on ADHD-200 \cite{bellec2017neuro} and ABIDE \cite{di2014autism} datasets. Using official preprocessing and quality control, ADHD-200 yields 776 subjects with 285 ADHD and 491 controls, and ABIDE yields 871 subjects with 403 ASD and 468 controls. 

Brain-HGCN comprises 3 attention layers ($d=64$) with learnable per-layer curvature $K_\ell$ (init=1.0) and 4 attention heads. We build sparse graphs ($k=10$) and compute the Fréchet mean over 5 iterations ($\eta=0.1$). The model was trained for 100 epochs using AdamW (lr=$10^{-3}$, weight decay=$5 \times 10^{-4}$, batch size=32).

\subsection{Comparsion with SOTA Methods}
As shown in Table \ref{tab1}, we compare Brain-HGCN with thirteen advanced baselines across CNN, Transformer, and GNN categories.

On ADHD-200, Brain-HGCN reaches accuracy 83.6\% and AUC 90.7\%, gains of 1.9 and 2.0 points over MM-GTUNets at 81.7\% and 88.7\%, with sensitivity 80.8\% and specificity 85.8\% forming the best combination. On ABIDE, Brain-HGCN attains accuracy 87.3\% and AUC 91.4\%, where AUC surpasses MM-GTUNets 87.4\% by 4.0 points and accuracy surpasses RTGMFF 86.4\% by 1.9 points; sensitivity 89.6\% and specificity 84.3\% are the highest overall. Across datasets, Brain-HGCN leads on core metrics and shows stronger cross-site robustness and discriminative power.

\subsection{Ablation Study}
Our ablation study (Table~\ref{tab:ablation}) quantifies each module's contribution and confirms its cross-cohort robustness on the two datasets.

\textbf{Hyperbolic geometry.}  
Removing hyperbolic geometry for a Euclidean one causes the largest performance drop, proving its critical role in modeling hierarchical brain networks with low distortion.

\textbf{Fr\'echet mean readout.}  
Removing the intrinsic Fr\'echet mean and averaging at a fixed base point consistently degrades performance, indicating that tangent-space pooling at the intrinsic mean eliminates base-point bias and stabilizes graph-level representations.

\textbf{Lorentzian attention.}  
Switching to Euclidean dot-product or removing attention causes further declines, highlighting the value of a geometry-consistent similarity metric.

\textbf{Signed aggregation.}  
Removing the signed aggregation mechanism reduces accuracy, which confirms the importance of distinguishing between excitatory and inhibitory pathways.

Overall, conclusions are consistent across datasets, the components are complementary, and the full model achieves the best results on key metrics, driven by its geometric and structural design.

\section{Conclusion}
In this study, we introduce Brain-HGCN, a hyperbolic graph convolutional network for brain functional network analysis. Brain-HGCN leverages the Lorentz model with geometry-consistent operations to reduce hierarchical distortion, incorporates a Lorentzian multi-head attention with signed aggregation to distinctly model positive and negative couplings, and employs an intrinsic Fréchet-mean readout to enable unbiased graph-level pooling. Comparative experiments on ADHD-200 and ABIDE demonstrate superior performance over strong baselines, alongside stable training and cross-site robustness, indicating its suitability for large-scale computational psychiatry studies. Future work will extend Brain-HGCN to dynamic functional connectivity and multimodal integration, explore site harmonization and domain generalization, and develop curvature-aware, interpretable biomarkers to facilitate clinical translation.



\bibliographystyle{IEEEbib}
\bibliography{refs}

\end{document}